\documentclass[10pt,twocolumn,letterpaper]{article}
\usepackage{cvpr}
\usepackage{times}
\usepackage{epsfig}
\usepackage{graphicx}
\usepackage{amsmath}
\usepackage{amssymb}
\usepackage{multirow, array}
\usepackage[utf8]{inputenc}
\usepackage{booktabs} 
\usepackage{makecell}
\usepackage{fontawesome}
\usepackage{pifont}
\usepackage{enumitem} 
\usepackage{tikz}
\newcommand\copyrighttext{%
\footnotesize \textcopyright 2025 IEEE. Personal use of this material is permitted.
Permission from IEEE must be obtained for all other uses, in any current or future
media, including reprinting/republishing this material for advertising or promotional
purposes, creating new collective works, for resale or redistribution to servers or
lists, or reuse of any copyrighted component of this work in other works.
}
\newcommand\copyrightnotice{%
\begin{tikzpicture}[remember picture,overlay]
\node[anchor=south,yshift=10pt] at (current page.south) {\fbox{\parbox{\dimexpr\textwidth-\fboxsep-\fboxrule\relax}{\copyrighttext}}};
\end{tikzpicture}%
}

\begin{document}

\title{Deep Data Hiding for ICAO-Compliant Face Images: A Survey}


\author{Jefferson David Rodriguez Chivata\\
University of Cagliari\\
Piazza d'Armi I - 09123 Cagliari (Italy)\\
{\tt\small jeffersond.rodriguez@unica.it}
\and
Davide Ghiani\\
University of Cagliari\\
Piazza d'Armi I - 09123 Cagliari (Italy)\\
{\tt\small davide.ghiani@unica.it}
\and
Simone Maurizio La Cava\\
University of Cagliari\\
Piazza d'Armi I - 09123 Cagliari (Italy)\\
{\tt\small simonem.lac@unica.it}
\and
Jefferson David Rodriguez Chivata\\
University of Cagliari\\
Piazza d'Armi I - 09123 Cagliari (Italy)\\
{\tt\small jeffersond.rodriguez@unica.it}
\and
Marco Micheletto\\
University of Cagliari\\
Piazza d'Armi I - 09123 Cagliari (Italy)\\
{\tt\small marco.micheletto@unica.it}
\and
Giulia Orr\'u\\
University of Cagliari\\
Piazza d'Armi I - 09123 Cagliari (Italy)\\
{\tt\small giulia.orru@unica.it}
\and
Federico Lama\\
Dedem S.p.A.\\
Via Cancelleria 59 - 00072 Ariccia (Italy)\\
{\tt\small federico.lama@dedem.it}
\and
Gian Luca Marcialis\\
University of Cagliari\\
Piazza d'Armi I - 09123 Cagliari (Italy)\\
{\tt\small marcialis@unica.it}
}

\maketitle

\copyrightnotice
\thispagestyle{empty}

\begin{abstract}
\vspace{-1em}
   ICAO-compliant facial images, initially designed for secure biometric passports, are increasingly becoming central to identity verification in a wide range of application contexts, including border control, digital travel credentials, and financial services. While their standardization enables global interoperability, it also facilitates practices such as morphing and deepfakes, which can be exploited for harmful purposes like identity theft and illegal sharing of identity documents. Traditional countermeasures like Presentation Attack Detection (PAD) are limited to real-time capture and offer no post-capture protection. 
   This survey paper investigates digital watermarking and steganography as complementary solutions that embed tamper-evident signals directly into the image, enabling persistent verification without compromising ICAO compliance. We provide the first comprehensive analysis of state-of-the-art techniques to evaluate the potential and drawbacks of the underlying approaches concerning the applications involving ICAO-compliant images and their suitability under standard constraints.
   We highlight key trade-offs, offering guidance for secure deployment in real-world identity systems.
\end{abstract}


\section{Introduction}

Biometric face recognition is central to identity management systems, particularly when secure and standardized verification is required. The International Civil Aviation Organization (ICAO) defines detailed specifications for facial image acquisition and formatting, which have been adopted globally in Machine Readable Travel Documents (MRTDs) such as biometric passports, and more recently in Digital Travel Credentials (DTCs) \cite{palaghias2024seamless,  wolf2018icao}. These standards are also increasingly used in remote identity verification systems, including financial services, where face images are employed to meet Know Your Customer (KYC) requirements \cite{hidayat2024face}.
While these standards ensure uniformity and interoperability, their predictable structure can be exploited to create manipulated yet compliant images through morphing \cite{di2024onot, ferrara2014magic}, generative methods \cite{yu2021survey}, and other image-based spoofing strategies \cite{erdogmus2014spoofing}. Furthermore, the long validity of ICAO images and the potential exfiltration through data breaches and public dissemination raise concerns about unauthorized reuse and biometric privacy \cite{laishram2025toward, prakasha2025privacy}.

Presentation Attack Detection (PAD) is a common defense against spoofing, but it operates only at capture time, without any protection once the image has been extracted, stored, or redistributed, and is limited by poor generalisation to novel attacks \cite{panzino2024evaluating,rusia2023comprehensive, shaheed2024deep}. Moreover, they often rely on additional sensors or computational modules, which may reduce throughput in real-time scenarios such as border control \cite{busch2024challenges, hernandez2023introduction}. To address these limitations, proactive data hiding techniques have been explored for encoding and embedding verifiable information directly into the image in a controlled and application-specific manner \cite{wang2023data}. These approaches aim to ensure integrity verification and traceability independently of the acquisition environment \cite{capasso2024comprehensive}.

Within this domain, digital watermarking and steganography represent two established paradigms \cite{wang2023data}. The former is primarily used to assert integrity or provenance, while the latter was traditionally developed to conceal auxiliary information for covert communication. Both techniques rely on modifying the visual signal to embed data that can be extracted later, typically without compromising the usability or appearance of the host image. In biometric contexts, these methods have been increasingly considered for embedding integrity markers or identifiers that remain robust under common transformations such as compression while preserving recognition performance and visual conformity with standard requirements \cite{sumalatha2024comprehensive, zhao2023proactive}.

Recent advances in deep learning have significantly improved data hiding methods \cite{wang2023data}, leveraging encoder-decoder networks \cite{zhu2018hidden}, generative adversarial networks (GANs) \cite{zhang2019robust}, transformers \cite{zhu2022stegformer}, invertible neural networks (INNs) and diffusion models \cite{xu2023stegdiff} to increase capacity, imperceptibility, and robustness. While some have been explored for biometric ID systems \cite{mcateer2019integration}, no survey has yet analyzed their applicability to ICAO-constrained biometric images. Existing surveys typically address watermarking and steganography from the perspective of general media security, copyright protection, or covert communication without considering the constraints imposed by biometric standards and operational requirements.

This survey fills that gap by systematically organizing and interpreting modern data-hiding approaches for post-acquisition certification under ICAO constraints.  Rather than replicating existing empirical results, we extract and recontextualize key insights from prior work, enabling a new comparative understanding of the strengths, limitations, and suitability of current methods in biometric certification scenarios.  Specifically, we provide: (i) a taxonomy of data hiding approaches categorized by robustness, architecture, and learning paradigm and their analysis in function of the risks associated with the real-world applications of ICAO-compliant images; (ii) a comparative analysis of recent watermarking and steganographic models based on such application scenarios; (iii) a critical discussion of their applicability for tamper detection and their potential role for ICAO scenarios.
Although the focus is on ICAO-compliant images, the insights extend to other facial recognition scenarios, including, but not limited to, those based on different imagery standards that impose comparable constraints. 
This analysis provides practical recommendations for designing secure systems by demonstrating that only a subset of existing deep learning techniques meets the combined requirements of imperceptibility, selective robustness, and reliable blind extraction needed for ICAO certification.

The rest of this manuscript is structured as follows. Section \ref{sec:ICAO} revises the key biometric and technical specifications of ICAO-compliant facial images and provides an overview of their real-world applications, as well as the potential threats and existing proactive countermeasures. Section \ref{sec:background} outlines the problem formulation, terminology, and key properties of data hiding techniques to certify biometric images. Section \ref{sec:modern} highlights the limitations of traditional data-hiding methods and explores modern learning-based approaches for watermarking and steganography in biometric image certification. Section \ref{sec:discussion} compares the underlying models, analyzing their potentialities and suitability in real-world applications. Finally, conclusions and future directions are presented in Section \ref{sec:conclusions}.


\section{ICAO Standards and Associated Threats}
\begin{table}[!t]
\centering
\caption{Summary of relevant ICAO Portrait Quality parameters for facial images in MRTDs and DTCs \cite{wolf2018icao}.}
\label{tab:icao_requirements}
\resizebox{\columnwidth}{!}{\begin{tabular}{lll}
\hline
\textbf{Parameter} & \textbf{Requirement} & \textbf{Section in \cite{wolf2018icao}} \\
\hline
Inter-eye distance 
& $\geq$ 90 px (recommended 120 px) & 5.2.4
\\
Background & Uniform, light-colored, without patterns & 5.2.5 \\
Lighting & Uniform illumination, no strong shadows & 5.2.6 \\
Pose & Full frontal, head vertically aligned & 5.3.1 \\
Facial expression & Neutral expression, mouth closed & 5.3.2 \\
Saturation (printed) & Non-background pixels with values 0 or 255 each $<$0.1\% & 6.3 \\
Resolution 
& 35 mm $\times$ 45 mm, scanned portrait $\geq$ 300 dpi & 6.4 \\
Compression format & JPEG (printed), JPEG2000 (logical storage) & 6.5 \\
\hline
\end{tabular}}
\end{table}
\label{sec:ICAO}

This section outlines the core biometric and technical specifications of ICAO-compliant facial images, reviews their deployment across real-world application domains, and examines associated manipulation threats and current proactive countermeasures.

\subsection{ICAO Requirements for Facial Image Quality}

Standardized facial image acquisition and encoding are fundamental to ensure interoperability and security in international identity verification systems. To this end, ICAO determines facial images as the primary biometric in MRTDs and DTCs, as detailed in Document 9303, Part 9 \cite{passports2021part}. The standard focuses on organizing biometric data within the Logical Data Structure (LDS) and mandates conformance to ISO/IEC 19794-5 \cite{iso19794}, later refined by ISO/IEC 39794-5 \cite{iso39794}, for the encoding of facial image data.

However, ICAO Document 9303 does not prescribe specific quality criteria for image acquisition, such as resolution, compression format, or subject pose. These operational aspects are addressed separately in the ICAO Technical Report on Portrait Quality (Version 1.0, 2018) \cite{wolf2018icao}, which provides best practice guidelines to ensure that captured portraits meet the operational needs of both automated and human identity verification processes.
Table \ref{tab:icao_requirements} summarizes the key parameters relevant to ICAO-compliant facial images. Within the ICAO documentation, the normative strength of each requirement is explicitly indicated: ``shall'' designates binding obligations, ``should'' denotes recommended best practices, and ``may'' identifies optional elements. This structured terminology balances technical consistency enforcement and operational flexibility.

While the standardization improves interoperability and comparison performance, it also introduces a highly predictable acquisition model. Adversaries can exploit the known constraints to synthesize or manipulate facial images that formally satisfy compliance checks, thereby increasing the difficulty of detecting fraudulent identities in critical verification workflows. The implications of these vulnerabilities and their impact on real-world identity systems are analyzed in the next section.

\subsection{Real-World Applications and Associated Risks}

ICAO-compliant images, originally developed for passport standardization, are now fundamental in governmental, financial, and commercial identity systems. In the travel sector, ICAO-compliant portraits enable Automated Border Control (ABC) through facial recognition using the biometric template stored in their electronic passports \cite{hidayat2024face}. DTCs are a digital extension of passports, allowing travelers to store their identity on mobile devices \cite{netherlands2023dtc, finnishborderguard2024dtc}, while still requiring ICAO-compliant facial images for global interoperability \cite{radutoiu2024study}. 
Beyond aviation, ICAO-compliant images are widely used in finance for biometric verification in KYC and onboarding, ensuring interoperability and accuracy even in remote authentication \cite{hayata2024trust, tavares2018wallid}. In addition, mobile identity systems and digital wallets increasingly rely on ICAO portraits to support secure user verification in diverse application ecosystems.

Although the operational benefits of standardization are evident, the predictability and stability of ICAO-constrained formats increase the exposure to several attack vectors. In addition to typical concerns associated with facial biometrics, such as aging effects and privacy risks \cite{guo2019survey, la20233d}, identity verification systems are increasingly challenged by sophisticated presentation attacks \cite{di2024onot}. For instance, morphing attacks, in which facial features from two individuals are blended to create a synthetic identity, can evade human inspection and automated recognition if compliance constraints are respected \cite{ferrara2014magic}. Similarly, deepfake techniques based on generative models can produce realistic ICAO images that embed fraudulent identities \cite{yu2021survey}. 

In parallel, the extension of ICAO standards beyond traditional passport systems further amplifies systemic exposure. In the context of DTCs, biometric data stored on mobile devices become vulnerable to compromise in device breaches \cite{prakasha2025privacy, sanna2024risk}. Similarly, financial institutions managing biometric databases for identity verification represent attractive targets for cyberattacks if proper security measures are not enforced \cite{laishram2025toward, tabassum2024blockchain}.  

Therefore, this combination of acquisition predictability, mass distribution, and extended operational validity underscores the need for protection mechanisms that persist beyond the moment of capture.

\subsection{Need for Proactive Protection Mechanisms}

PAD is the primary defense against biometric spoofing, analyzing cues like texture inconsistencies, motion artifacts, or liveness to detect falsified traits at the point of capture \cite{sharma2023survey}. Despite increasing sophistication, its protection is limited to acquisition event. Once the image is stored or transmitted, PAD offers no safeguard against tampering, synthetic alterations, or unauthorized redistribution \cite{busch2024challenges, shaheed2024deep}.

In application scenarios involving ICAO-compliant images, where biometric data may circulate across decentralized infrastructures and remain valid for extended periods, the absence of persistent integrity verification becomes a critical vulnerability. Traditional cryptographic techniques can secure the transmission but offer no guarantees once access to the content is obtained. Proactive security strategies have been proposed to address this gap by embedding verifiable integrity markers directly within the biometric content. Unlike capture-time defenses, embedded signals persist across the image lifecycle, enabling post-hoc verification of authenticity and tamper-evidence. Among the candidate approaches, data hiding techniques offer promising solutions to enhance the resilience of ICAO-compliant facial images without compromising their usability for visual inspection and automated recognition \cite{capasso2024comprehensive, sumalatha2024comprehensive}. Accordingly, the following section surveys the fundamental data-hiding methods, focusing on their design principles, embedding strategies, and relevance to biometric integrity protection.

\section{Data Hiding for Biometric Image Certification}
\label{sec:background}

This section provides the necessary background on data-hiding techniques for biometric image certification, presenting the problem formulation, key terminology, and fundamental properties that guide the following evaluation of such systems in ICAO-compliant contexts.

\begin{figure}[t]
    \centering
    \includegraphics[width=\columnwidth]{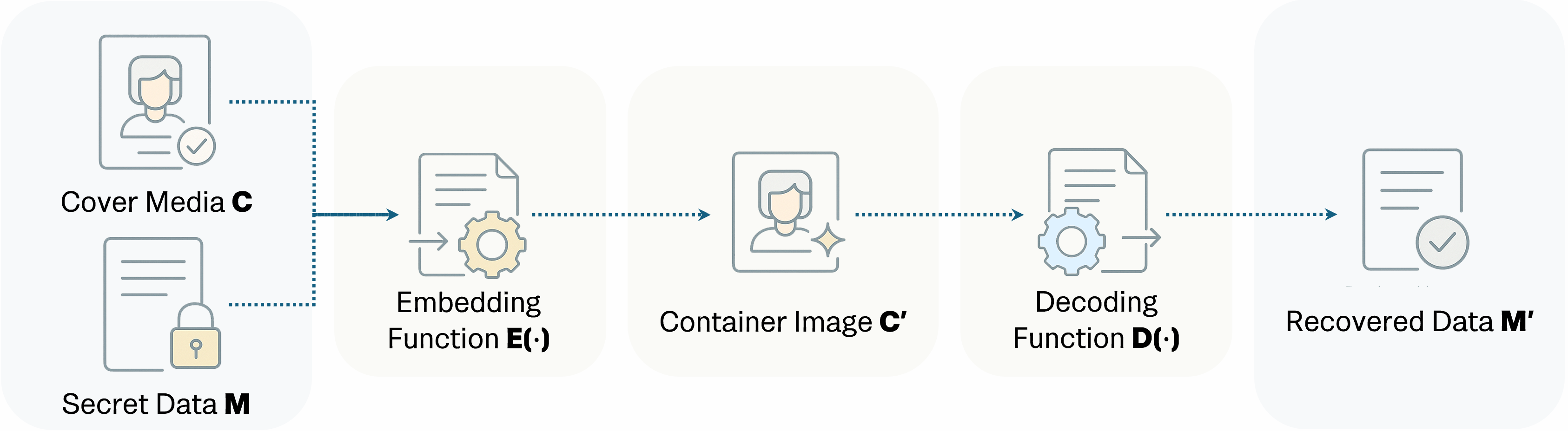}
    \caption{General scheme of a data hiding system.}
    \label{fig:threatmodel}
\end{figure}

\subsection{Problem formulation and terminology}

The certification of ICAO-compliant biometric images requires embedding security-relevant information directly into the image content, in a manner that preserves its operational usability for recognition and document issuance. In this framework, the data hiding process is modeled by two functions: embedding and extraction (Figure \ref{fig:threatmodel}). Given a cover image $\mathbf{C}$ and a secret message $\mathbf{M}$, the embedding function $E(\cdot)$ generates a container image $\mathbf{C'}$ according to:
\begin{equation}
\mathbf{C'} = E(\mathbf{C}, \mathbf{M}) 
\end{equation}
where $\mathbf{C'}$ should maintain a high degree of visual and biometric similarity to $\mathbf{C}$. The hidden message is subsequently recovered via a decoding function $D(\cdot)$, producing an estimate $\mathbf{\hat{M}}$:
\begin{equation}
\mathbf{\hat{M}} = D(\mathbf{C'})
\end{equation}

\begin{figure}[t]
    \centering
    \includegraphics[width=0.9\columnwidth]{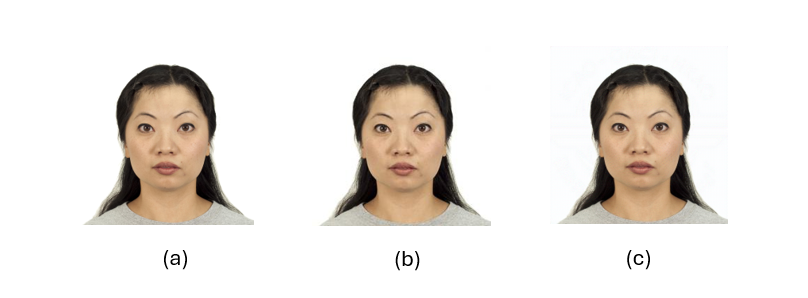}
    \caption{Comparison between: a) original input image; b) watermarked image obtained with \cite{yu2024effective} ($PSNR=45.813$, $SSIM=0.9861$ for 1 bpp) ; c) steganographic image obtained with \cite{ke2024stegformer} ($PSNR=39.562$, $SSIM=0.9699$ for 24 bpp).}
    \label{fig:examples}
\end{figure}

Throughout this work, we consider data hiding methods whose design and evaluation are driven by the specific needs of biometric certification, rather than by general-purpose communication or copyright protection scenarios.
Based on their operational goals, these methods can be categorized into two broad functional classes:
\begin{itemize}
    \item \textit{Digital watermarking}: these methods embed information to assert properties such as authenticity, integrity, or provenance of the cover image. The embedded data is expected to survive benign transformations and remain detectable, thus enabling certification even after typical processing such as compression or scaling.
    \item \textit{Steganography}: these methods aim to conceal the existence of the embedded information, maximizing imperceptibility and minimizing detectability. In the context of biometric certification, steganographic techniques can be reinterpreted to embed fragile integrity signals that, while remaining imperceptible, are disrupted by malicious modifications.
\end{itemize}

The choice between watermarking and steganography depends on the threat model and operational requirements. Watermarking is typically preferred when persistent verification across benign transformations is desired. Conversely, steganographic embedding may be advantageous when the primary goal is the detection of unauthorized alterations without introducing perceptible changes. 

\subsection{Key Properties of Data Hiding Systems}
\label{sec:dh_properties}
Several characteristics and structural properties define data hiding methods in biometric certification contexts, affecting usability, transparency, and extraction requirements. In the following, we describe the most relevant properties \cite{wang2023data}:
\textit{Embedding visibility}: Data hiding techniques can produce either visible or invisible embeddings. In visible embedding, the presence of the hidden information is perceptible to human observers, serving as an overt signal. Invisible embedding seeks to maintain the perceptual indistinguishability between $\mathbf{C}$ and $\mathbf{C'}$, minimizing the risk of detection.

\textit{Blind vs Non-Blind Extraction}: In blind data hiding systems, the decoding function $D(\cdot)$ operates exclusively on the container image $\mathbf{C'}$, requiring no access to the original cover $\mathbf{C}$ or any external auxiliary information. Formally: $\mathbf{\hat{M}} = D(\mathbf{C'})$. In non-blind systems, successful extraction depends on additional side information, typically the original cover image $\mathbf{C}$, leading to a decoding function of the form: $\mathbf{\hat{M}} = D(\mathbf{C'}, \mathbf{C})$.

\textit{Cover-Based vs Coverless Embedding}: Cover-based approaches start from a given cover image $\mathbf{C}$ and embed the message $\mathbf{M}$ to produce $\mathbf{C'}$. Coverless approaches, in contrast, generate $\mathbf{C'}$ directly conditioned on $\mathbf{M}$ without relying on an existing cover. 

\textit{Invertibility}: Invertible data hiding methods allow the simultaneous recovery of the embedded message $\mathbf{M}$ and, optionally, the original cover image $\mathbf{C}$ from the container $\mathbf{C'}$.

 \textit{Security}: A secure data hiding method must prevent unauthorized detection or extraction of the embedded data. Adversaries may attempt to detect the presence of embedded data via steganalysis techniques or to decode it without access to the original embedding process or keys.

\textit{Fragility and Robustness}: The resilience of the embedded message against transformations determines if a method is classified as fragile, semi-fragile, or robust. Fragile methods are highly sensitive to any alteration of the container image $\mathbf{C'}$, leading to significant degradation or loss of embedded information $\mathbf{M}$ even under minor modifications. Semi-fragile methods are designed to withstand benign operations like compression, but fail under malicious semantic manipulations, such as face morphing or swapping. Robust methods aim to maintain the integrity of the embedded data across a broad range of distortions, as signal processing and geometric transformations, and adversarial attacks, ensuring the persistence of hidden data under diverse operational conditions.

\textit{Capacity}: The capacity of a data-hiding system defines the maximum amount of information that can be embedded and reliably extracted from a container image $\mathbf{C'}$.

The relevance of these properties varies regarding the operational constraints of ICAO-compliant biometric image certification. In this domain, the invisibility of the embedding is mandatory to avoid degradation of recognition performance and comply with visual quality standards. Blind extraction is highly desirable to enable decentralized verification without requiring access to the original cover image. Cover-based embedding is mandatory to ensure the certification process refers to an authentic biometric acquisition. Invertibility is also desired to verify the integrity of the embedded message and the cover image after potential manipulations. Security against unauthorized detection or extraction is desirable to protect the confidentiality of embedded data. Fragile and semi-fragile embedding strategies are mandatory to enable reliable tamper detection, while robust embedding approaches are unsuitable as they may tolerate unacceptable semantic alterations. Finally, capacity must be balanced to carry necessary certification data without compromising invisibility or biometric performance: too little limits effectiveness, while too much may introduce artifacts due to stronger modifications in cover-based embedding.
Accordingly, the remainder of this survey focuses on data hiding methods that satisfy the operational requirements identified above. 
Figure~\ref{fig:examples} shows examples of cover and container image pairs, illustrating these combined requirements for ICAO-compliant systems.

\section{Fragile and Semi-Fragile Data Hiding}\label{sec:modern}

\subsection{Limitations of Traditional Methods}

Early fragile and semi-fragile data hiding systems predominantly relied on traditional embedding techniques operating in the spatial or frequency domain. Spatial domain approaches, such as Least Significant Bit (LSB) substitution or histogram modification \cite{chan2004hiding, coatrieux2012reversible}, offered good imperceptibility but were highly vulnerable to benign transformations like compression, filtering, or even minor noise. Frequency domain methods, relying on transformations such as DFT \cite{rawat2012blind}, DWT \cite{hamidi2018hybrid} or DCT \cite{hamidi2018hybrid}, improved robustness against certain signal processing operations, such as compression, but often suffered perceptual distortion and lacked fine-grained control over the fragility of embedded information \cite{cox1996secure}. Although these traditional algorithms have shown effectiveness in specific integrity verification tasks, their applicability is inherently narrow, requiring expert-driven design tailored to specific scenarios. Furthermore, the increasing sophistication of manipulation and removal attacks compromises their long-term reliability \cite{geng2020real}. 

The advent of deep learning introduced a paradigm shift in data hiding. Deep neural networks provide adaptable and generalized frameworks capable of learning complex embedding patterns directly from data. This enables improved resilience against a broader range of attacks, enhanced imperceptibility, and the possibility of dynamically retraining models to prioritize different objectives, such as robustness, invisibility, or payload capacity, without requiring specialized manual engineering \cite{abdelnabi2021adversarial, zhu2018hidden}. Moreover, the non-linearity of deep architectures significantly enhances the security of the embedded information against adversarial retrieval attempts.
Considering these advantages, learning-based methods have become a promising direction for developing fragile and semi-fragile data hiding systems suitable for modern biometric image certification under ICAO-compliant constraints. 

\subsection{Fragile and Semi-Fragile Watermarking}

Recent deep learning-based watermarking approaches have explored several architectural paradigms that can meet the operational demands of ICAO-compliant biometric certification. Those based on encoder–decoder, GANs, transformers, and INNs have been explored with varying degrees of success and constitute the predominant design strategies. Each architecture presents distinct trade-offs between invisibility, robustness to benign transformations, and sensitivity to semantic manipulations.

Encoder–decoder frameworks represent the foundational architecture adopted in most modern watermarking solutions due to their conceptual simplicity and flexibility. Typical implementations employ convolutional neural networks to embed a carefully controlled payload into cover images, optimizing embedding imperceptibility while preserving extraction accuracy.

A seminal method, \textit{HiDDeN}~\cite{zhu2018hidden}, introduced an end-to-end GAN framework integrating encoder, decoder, and discriminator, using adversarial training to embed resilient watermarks. Despite its innovation, \textit{HiDDeN} showed limited capacity and generalization against common signal attacks, prompting further developments. For instance, \textit{ARWGAN}~\cite{huang2023arwgan} incorporated attention-guided feature fusion and dense connections within a full GAN pipeline, enhancing both imperceptibility and robustness. Nonetheless, GAN-based methods still tend to introduce subtle artifacts during generation. Furthermore, their emphasis on synthesis rather than explicit recovery can limit watermark extraction reliability, even under known benign operations.

To mitigate these issues, some methods rely solely on CNN-based architectures, incorporating discriminators and adversarial training while minimizing generative components. For example, \textit{MBRS}~\cite{jia2021mbrs} proposed a robust end-to-end design against JPEG compression by simulating noise layers during training. Other works like \textit{TDSL}~\cite{liu2019novel} and \textit{Adaptor}~\cite{wang2023adaptor} employ a two-phase training and embedding strategy to improve resilience against realistic, non-differential transformations like JPEG compression. These methods also allow tunable embedding strength, providing a more flexible trade-off between imperceptibility and robustness. Beyond resilience to benign distortions, approaches such as \textit{FaceSigns}~\cite{neekhara2022facesigns} and \textit{WaterLo}~\cite{beuve2023waterlo} expanded their semi-fragile scopes by also including malicious semantic transformations such as face swap within the training process. This enables the watermark to remain robust to expected benign changes (e.g., JPEG compression or filtering) while becoming sensitive to semantic manipulations. 
Although CNN-based architectures with discriminators have demonstrated strong performance, recent transformer-based models have emerged, leveraging attention mechanisms to embed watermarks in spatially relevant regions adaptively. \textit{StegaFormer}~\cite{yu2024effective} and \textit{WFormer}~\cite{luo2024wformer} have shown significant gains in both imperceptibility and extraction accuracy, in exchange for higher computational requirements. Their performance in semi-fragile contexts suggests promising applicability to ICAO standards.
 
Finally, despite the versatility of encoder–decoder schemes and their synergy with adversarial training, they often suffer from information loss and require precise tuning between robustness and fragility. To address this, INN-based methods like \textit{RIS} \cite{lan2023robust} offer promising alternatives. Originally explored in steganography, INNs show great potential in ensuring total invertibility and higher imperceptibility, key factors for ICAO-compliant systems.

\subsection{Fragile and Semi-Fragile Steganography}

Initially developed for covert communication, steganographic methods can be reinterpreted in biometric image certification to act as fragile integrity markers \cite{ghiani2025fragile}. By embedding sensitive payloads designed to degrade under tampering, steganography offers a complementary strategy for proactively detecting unauthorized modifications in ICAO-compliant scenarios. As in watermarking systems, deep learning-based steganographic methods adopt a variety of architectural paradigms, including encoder–decoder frameworks, GANs, INNs, transformers, and diffusion models.

Encoder–decoder architectures based on fully DNNs and CNNs \cite{baluja2017hiding,zhang2020udh} provide a standard framework for fragile and semi-fragile steganography, embedding entire images (high capacity) into host content while optimizing invisibility and reconstruction accuracy. These systems, despite targeting high visual fidelity, inherently exhibit sensitivity to content alterations, making them suitable for tamper detection. Variants that incorporate additional embedding constraints, such as symmetry preservation \cite{khalifa2022imperceptible}, further refine the balance between imperceptibility and fragility, strengthening the potential of encoder-decoder architectures for integrity verification tasks.

GAN-based steganography enhances invisibility and undetectability through adversarial training. Early methods like \textit{SteganoGAN} \cite{zhang2019steganogan} and \textit{ISGAN} \cite{zhang2019invisible} offer good capacity and visual quality but lack robustness to perturbations such as compression. Recent models, including \textit{ADBH} \cite{yu2020attention}, \textit{CHAT-GAN} \cite{tan2021channel}, and \textit{Cover-GAN} \cite{li2024cover}, address this by incorporating attention mechanisms or perturbation simulation, improving robustness. However, they still face decoding challenges under lossy conditions and limited scalability to high-resolution images \cite{zhang2019steganogan, zhang2019invisible, yu2020attention, tan2021channel, li2024cover}.

INNs have gained attention in steganography for their ability to model concealing and revealing as symmetric, reversible processes \cite{jing2021hinet,lu2021large}. These architectures support high capacity and imperceptibility, often outperforming traditional encoder-decoder schemes in preserving image fidelity. Recent works have introduced explicit robustness mechanisms, such as conditional flows \cite{xu2022robust} or direct embedding in DCT coefficients \cite{lan2023robust} to improve robustness under distortions like JPEG compression. While computational demands and invertibility constraints can limit scalability, their capacity to ensure accurate decoding even under moderate image degradation makes them a promising option for scenarios requiring integrity and resilience, such as biometric image certification under ICAO standards.

Transformer-based steganography represents an emerging direction. Models such as \textit{Stegformer} \cite{ke2024stegformer} exploit attention mechanisms to distribute the payload across semantically meaningful regions adaptively. This flexibility could facilitate embedding strategies focused on critical biometric features to enhance tamper sensitivity. However, these transformer-based models still prioritize payload capacity and general robustness. Specific adaptations may be necessary to align with fragile or semi-fragile requirements.

Finally, diffusion models have recently been explored for generative steganography \cite{yu2023cross,yang2024diffstega}, synthesizing entire images conditioned on hidden messages. Although techniques like \textit{DERO} \cite{fang2024dero} achieve state-of-the-art imperceptibility and steganalysis resistance, their generative nature makes them unsuitable for certifying the authenticity of pre-acquired biometric images, as required by ICAO standards.

\section{Comparative Evaluation}
\label{sec:discussion}

The evaluation of fragile and semi-fragile data hiding methods is structured along four core dimensions: imperceptibility, robustness, capacity, and security.
While a broader set of structural properties has been introduced in Section \ref{sec:dh_properties} to characterize the behavior of data hiding systems, this section focuses on the quantitative metrics associated with the most critical dimensions for ICAO-compliant biometric certification. Accordingly, each metric must be interpreted in light of the operational objectives of proactive tamper detection. The remainder of this section formally introduces the evaluation criteria and presents a comparative analysis of representative methods along these dimensions.

\subsection{Evaluation Metrics}
\textbf{Imperceptibility} The visual fidelity between the original image $\mathbf{C}$ and the container image $\mathbf{C'}$ is critical for ICAO-compliant biometric images, where any visible alteration may compromise recognition performance. Standard evaluation metrics include Peak Signal-to-Noise Ratio (PSNR) \cite{hore2010image}, Structural Similarity Index (SSIM) \cite{hore2010image}, learning-based perceptual metrics such as LPIPS \cite{zhang2018unreasonable}, and Mean Square Error (MSE). Although high imperceptibility is essential across all robustness levels, it becomes particularly critical for fragile and semi-fragile methods targeting biometric certification scenarios \cite{nadimpalli2024social,neekhara2022facesigns}.

\textbf{Robustness} The evaluation of the ability of a data hiding method to preserve and accurately recover the embedded message $\mathbf{M}$ from the container image $\mathbf{C'}$ after undergoing various transformations must consider both benign operations, such as JPEG compression or resizing \cite{wang2022staged,wu2023sepmark}, and more severe and malicious alterations, such as adversarial perturbations \cite{huang2022cmua} or semantic manipulations \cite{nadimpalli2024social,wu2023sepmark}.
The primary metric for evaluating robustness is the Bit Error Rate (BER), which quantifies the proportion of bits incorrectly recovered between $\mathbf{M}$ and $\mathbf{\hat{M}}$. Lower BER values correspond to greater resilience of the hidden information under distortions. In some cases, Bit Recovery Accuracy (BRA) or normalized cross-correlation (NC)~\cite{zhao2023proactive} are also used to assess the fidelity of message retrieval, particularly under different types of attacks.
For fragile and semi-fragile data hiding tailored to ICAO-compliant biometric certification, robustness must be properly tuned: the embedded payload should resist benign signal-level degradations, specifically JPEG compression, but fail when semantic integrity is compromised \cite{neekhara2022facesigns,beuve2023waterlo}. Excessive robustness, as in copyright watermarking frameworks \cite{wang2024invisible,zhang2024editguard}, would undermine the ability to detect unauthorized biometric modifications.

\textbf{Capacity}  
The amount of information that can be embedded within an image is typically measured in bits per pixel (BPP)~\cite{cox2007digital}. This metric is particularly used in steganographic approaches, where the objective is to covertly transmit large volumes of data without arousing suspicion. High-capacity methods enable the embedding of complex payloads such as multiple images~\cite{guan2022deepmih,ke2024stegformer,lu2021large}, encryption keys, or metadata, but often at the expense of visual fidelity.
In contrast, capacity could not be the primary goal in watermarking contexts. However, maintaining a reasonable embedding capacity can offer operational flexibility, for instance, by allowing individualized watermarking across users or systems, provided that imperceptibility and fragility requirements are not violated.

\textbf{Security}  
The resistance of a data hiding system against intentional attacks aiming to detect, remove, or corrupt the embedded message $\mathbf{M}$ can be measured in different ways. Evaluation typically includes resilience against adversarial perturbations designed to disrupt decoding \cite{zhang2024dual}, detection by steganalysis models \cite{hu2023invisible}, watermark removal techniques \cite{nadimpalli2024social}, and generalization to manipulations not encountered during training, such as deepfake generation \cite{wang2024lampmark} or presentation attacks \cite{guo2024coverless}. Common metrics include bit recovery accuracy after an attack, steganalysis error rates, attack success rates \cite{wang2024invisible}, and deepfake detection performance \cite{zhao2023proactive}. In deep learning-based systems, evaluations are often conducted under white-box (i.e., the attacker knows the model) and black-box (i.e., model unknown) scenarios to comprehensively assess vulnerability \cite{jiang2024certifiably, wang2021faketagger}. In ICAO-compliant applications, security is crucial to ensure that the embedded data resists unauthorized modifications while remaining imperceptible and non-disruptive to recognition systems.

\subsection{Comparative Analysis and Discussion}

\begin{table*}[ht]
  \centering
  \small
      \caption{
Deep data hiding methods comparison across imperceptibility, robustness, and payload. Underlined values indicate PSNR between cover/stego images; italicized values refer to PSNR between secret/recovered images. BER denotes the bit error rate (\%) between embedded/extracted messages (before/after) JPEG compression. Asterisk (*) indicates values under JPEG compression with QF=50.
  }

  \resizebox{\textwidth}{!}{%
    \begin{tabular}{|c|c|c|c|c|c|c|c|c|c|c|c|}
      \hline
      \multirow{2}{*}{\textbf{Model}} 
      & \multirow{2}{*}{\textbf{\makecell{Venue\\Year}}} & \multirow{2}{*}{\textbf{\makecell{Framework\\Type}}} 
      & \multirow{2}{*}{\textbf{Domain}} 
      & \multirow{2}{*}{\textbf{Input}} 
      & \multirow{2}{*}{\textbf{\makecell{Dataset/\\Image dimension}}} 
      & \multirow{2}{*}{\textbf{\makecell{Payload\\(BPP)}}} 
      & \multirow{2}{*}{\textbf{\makecell{Imperceptibility\\(PSNR dB)}}} 
      & \multicolumn{2}{c|}{\makecell[c]{\textbf{JPG Compression Robustness} \\ \textbf{(QF=90,*=50)}}}
      & \multirow{2}{*}{\textbf{\makecell{Extra\\ robustness}}} 
      & \multirow{2}{*}{\textbf{Grade}} \\
      \cline{9-10}
       & & & & & & & & \textbf{PSNR (dB)} & \textbf{BER (\%)} & & \\
      \hline
      \makecell[c]{SteganoGAN \\ \cite{zhang2019steganogan}} & \makecell[c]{CoRR \\ 2019} & \makecell{Enc-Dec\\ (GAN)} & Spatial & Data Hiding & \makecell[c]{\textbf{DIV2K}/\,$\sim1024\times1024$ \\ \textbf{COCO}/\,$\sim256\times256$} & $6$ 
      & \makecell[c]{$\underline{38.94}$ / $-$ \\ $\underline{36.33}$ / $-$} & $-$ & $-$ & \ding{55}  & Low+ \\
      \hline
      ISGAN \cite{zhang2019invisible} & \makecell[c]{MTAP \\ 2019} & \makecell{Enc-Dec\\ (GAN)} & Spatial & Image Hiding
        & \makecell[c]{\textbf{LFW}/ $256\times256$ \\ \textbf{PASCALVOC}/ $256\times256$ \\ \textbf{ImageNet}/ $256\times256$} & $8$
        & \makecell[c]{\underline{34.63} / \textit{33.63} \\ \underline{34.49} / \textit{33.31} \\ \underline{34.89} / \textit{33.42}} & $-$ & $-$ & \ding{55}  & Low \\
      \hline
      ABDH \cite{yu2020attention} & \makecell[c]{AAAI \\ 2020} & \makecell{Enc-Dec\\ (GAN)} & Spatial & Image Hiding & \textbf{COCO}/\,$512\times512$ & $\sim24$
        & \underline{31.91} / \textit{30.66} & $-$ / \textit{32.97} & $-$ & \checkmark & Low \\
      \hline
      UDH \cite{zhang2020udh} & \makecell[c]{NeurIPS \\ 2020} & \makecell{Enc-Dec\\ (CNN)} & Spatial & Image Hiding & \textbf{ImageNet}/\,$128\times128$ & $\sim24$
        & \underline{39.13} / \textit{35.0} & $-$ & \textit{0.0} / \textit{0.6*} & \checkmark & Medium \\
      \hline
      MBRS \cite{jia2021mbrs} & \makecell[c]{ACM-MM \\ 2021} & \makecell{Enc-Dec\\ (CNN)} & Spatial & \makecell[c]{Binary String \\ Hiding}
        & \makecell[c]{\textbf{ImageNet}/ $400\times400$ \\ \textbf{COCO}/ $\sim400\times400$}
        & $\sim 0.0039$ & \makecell[c]{$-$ \\ \underline{39.32} / $-$} 
        & \makecell[c]{$-$ \\ \underline{42.04} / $-$} 
        & \makecell[c]{$-$ \\ \textit{0.0012} / \textit{0.00063}} & \ding{55}  & High \\
      \hline
      ISN \cite{lu2021large} & \makecell[c]{CVPR \\ 2021} & INN & Spatial & Image Hiding
        & \makecell[c]{\textbf{ImageNet}/ $144\times144$ \\ \textbf{Paris Street}/ $144\times144$} & $\sim24$
        & \makecell[c]{\underline{38.05} / \textit{35.38} \\ \underline{40.49} / \textit{43.33}} & $-$ & $-$ & \ding{55}  & Low+ \\
      \hline
      HiNet \cite{jing2021hinet} & \makecell[c]{CVPR \\ 2021} & INN & \makecell{Frequency\\(DWT)} & Image Hiding
        & \makecell[c]{\textbf{DIV2K}/ $1024\times1024$ \\ \textbf{ImageNet}/ $256\times256$ \\ \textbf{COCO}/ $256\times256$} & $\sim24$
        & \makecell[c]{\underline{48.99} / \textit{52.86} \\ \underline{44.60} / \textit{46.78} \\ \underline{46.52} / \textit{46.98}} & $-$ & $-$ & \ding{55} & Medium+ \\
      \hline
      \makecell[c]{FaceSigns \\ \cite{neekhara2022facesigns}} & \makecell[c]{TOMM \\ 2022} & \makecell{Enc-Dec\\ (CNN)} & Spatial & \makecell[c]{Binary String \\ Hiding}
        & \textbf{CelebA}/\,$256\times256$
        & $0.00065$ & \underline{36.08} / $-$ & $-$ & \textit{0.32} / \textit{0.51} & \checkmark & Medium- \\
      \hline
      RIIS \cite{xu2022robust} & \makecell[c]{CVPR \\ 2022} & INN & Spatial & Image Hiding
        & \makecell[c]{\textbf{DIV2K}/ $1024\times1024$ \\ \textbf{ImageNet}/ $\sim256\times256$}
        & $\sim24$ & \makecell[c]{$-$ / \textit{44.19} \\ \underline{43.97} / \textit{46.71}} 
        & \makecell[c]{$-$ / \textit{28.71} \\ \underline{28.17} / \textit{28.53}} & $-$ & \checkmark & Medium+ \\
      \hline
      RIS \cite{lan2023robust} & \makecell[c]{AAAI \\ 2023} & INN & \makecell{Frequency\\(DCT)} & \makecell[c]{Binary String \\ Hiding}
        & \textbf{MSCOCO}/\,$256\times256$
        & $1$ & \underline{48.41} / $-$ & \underline{44.13} / $-$ & \textit{0.0} / \textit{0.31} & \ding{55}  & High+ \\
      \hline
      \makecell[c]{DeepMIH \\ \cite{guan2022deepmih}} & \makecell[c]{PAMI \\ 2022} & INN & \makecell{Frequency\\(DWT)} & Image Hiding
        & \makecell[c]{\textbf{DIV2K}/ $1024\times1024$ \\ \textbf{ImageNet}/ $256\times256$ \\ \textbf{COCO}/ $256\times256$}
        & $\sim24$
        & \makecell[c]{\underline{43.72} / \textit{41.41} \\ \underline{40.31} / \textit{36.63} \\ \underline{40.30} / \textit{36.55}} & $-$ & $-$ & \ding{55}  & Medium+ \\
      \hline
      Adaptor \cite{wang2023adaptor} & \makecell[c]{TCSVT \\ 2023} & \makecell{Enc-Dec\\ (CNN)} & Spatial & \makecell[c]{Binary String \\ Hiding} 
        & \textbf{COCO}/\,$128\times128$ & $\sim 0.0013$
        & $-$ & \underline{38.42*} / \textit{$-$} & $-$ / \textit{0.016*} & \checkmark & Medium- \\
      \hline
      \makecell[c]{DIH-OAIN \\ \cite{hu2023invisible}} & \makecell[c]{TCSVT \\ 2023} & INN & Spatial & Image Hiding
        & \makecell[c]{\textbf{COCO}/ $\sim256\times256$ \\ \textbf{PASCALVOC}/ $\sim256\times256$}
        & $\sim24$
        & \makecell[c]{\underline{46.56}/ \textit{39.73} \\ \underline{54.45}/ \textit{48.60}} & $-$ & $-$ & \ding{55}  & Medium- \\
      \hline
      \makecell[c]{StegFormer \\ \cite{ke2024stegformer}} & \makecell[c]{AAAI \\ 2024} & \makecell{Enc-Dec\\ (Transformer)} & Spatial & Image Hiding
        & \makecell[c]{\textbf{DIV2K}/ $1024\times1024$ \\ \textbf{ImageNet}/ $256\times256$ \\ \textbf{COCO}/ $256\times256$} & $\sim24$
        & \makecell[c]{\underline{56.30} / \textit{55.45} \\ \underline{48.79} / \textit{49.18} \\ \underline{48.77} / \textit{49.21}} & $-$ & $-$ & \ding{55} & Medium+ \\
      \hline
      \makecell[c]{Stegaformer \\ \cite{yu2024effective}} & \makecell[c]{BMVA \\ 2024} & \makecell{Enc-Dec\\ (Transformer)} & Spatial & \makecell[c]{Binary String \\ Hiding}
        & \makecell[c]{\textbf{COCO}/ $256\times256$ \\ \textbf{DIV2K}/ $256\times256$}
        & $3$ & \makecell[c]{\underline{43.37} / $-$ \\ \underline{47.31} / $-$} 
        & $-$ & \makecell[c]{\textit{0.32} / $-$ \\ \textit{0.24} / $-$} & \ding{55}  & Medium- \\
      \hline
    \end{tabular}%
  }
  \label{tab:bigcomparison}
\end{table*}

Table \ref{tab:bigcomparison} offers a comparative overview of state-of-the-art deep learning-based data hiding approaches assessed across core properties relevant to biometric image certification: imperceptibility, robustness, and payload capacity. To ensure fair comparison, all methods are analyzed using metrics reported in their original publications, also considering their applicability to ICAO-compliant scenarios. For payload capacity, a similar value was used to facilitate comparison rather than reporting the maximum capacity. The table also reports their architecture (e.g., INNs, transformers), the embedding domain (spatial or frequency), and the hidden content type.

For each method, imperceptibility is quantified using the PSNR between the original image ($C$) and the container ($C'$). Values above 40 dB generally denote visually indistinguishable changes \cite{sadek2017robust, sun2022high}, crucial in the context of ICAO-compliant facial images, where any visible degradation may affect both human inspection and automated face recognition. Several methods, such as \textit{HiNet}, \textit{RIIS}, \textit{RIS}, and \textit{StegFormer}, consistently exceed this threshold across multiple datasets and resolutions, indicating a strong alignment with ICAO visual quality requirements.

In parallel, recovery fidelity, measuring how accurately the hidden data can be retrieved, is reported as PSNR between secret and recovered images or, in the case of binary string hiding, as the BER. High reconstruction PSNRs ($\geq$35 dB) and low BERs ($\leq$0.3\%) are indicative of practical message preservation under distortion-free conditions (e.g., \cite{ghiani2025fragile}). Approaches such as \textit{StegFormer}, \textit{MBRS}, \textit{RIS}, and \textit{FaceSigns} demonstrate excellent recovery performance, suggesting strong potential for scenarios in which embedded information must be reliably extracted post-verification.

Robustness to JPEG compression is reported as PSNR degradation or BER under compression with quality factors (QF) of 90 or 50. This form of selective robustness is essential for semi-fragile watermarking scenarios. In this context, the ``\textit{Extra robustness}'' column indicates whether a method also preserves the payload under other operations, such as filtering or geometric changes. While general-purpose watermarking methods typically aim for strong resilience, such robustness may be counterproductive in biometric certification, where fragility to malicious content manipulation is a desired property. Accordingly, methods that are not robust to arbitrary transformations are better aligned with the ICAO-compliant image integrity verification requirements.

Finally, the overall ``\textit{Grade}'' column provides a high-level qualitative indication of each method's potential suitability in ICAO contexts, reflecting a balanced assessment across imperceptibility, recovery fidelity, robustness behavior, and compliance with minimum image resolution requirements (Table \ref{tab:icao_requirements}). While some approaches, like \textit{SteganoGAN} or \textit{ISGAN}, exhibit good imperceptibility, their lack of robustness and recovery fidelity limits their applicability. Conversely, methods such as \textit{RIS} and \textit{MBRS} strike a more favorable trade-off, suggesting higher compatibility with the goals of biometric image certification. Importantly, INN-based and transformer-based designs appear particularly promising due to their inherent support for invertibility and flexible embedding.
While the security property is a critical dimension in our analysis, a detailed comparative assessment is challenging. This is because most of the suitable methods report performance against generic steganalysis benchmarks (e.g., with detectability rates around 55\% or less using methods like XuNet~\cite{xu2016structural} or SRNet~\cite{boroumand2018deep}), which may not reliably reflect the security required against sophisticated, context-aware attacks in real-world ICAO operational environments.
Despite the table's qualitative nature, which limits strict quantitative ranking based on application priorities, it represents the first attempt to systematically assess the suitability of data-hiding approaches for ICAO-compliant scenarios. Specifically, reported values across key properties offer practitioners a valuable basis for selecting appropriate methods for specific operational needs. Notably, the analysis reveals that only a subset of models meet the combined requirements of visual conformity, selective robustness, and reliable decoding necessary for biometric image certification. These findings underscore the need for further development of specialized strategies tailored to biometric identity constraints.

\section{Conclusions and Future Directions}
\label{sec:conclusions}

This survey explored how steganographic and watermarking techniques can enhance the integrity and traceability of ICAO-compliant facial images, especially where traditional countermeasures like PAD offer limited post-capture protection. These methods are particularly relevant in high-risk scenarios such as border control and KYC, where manipulation attacks like morphing and deepfake attacks pose serious threats. 
We reviewed core concepts, categorized existing techniques, and analyzed their suitability under ICAO constraints, emphasizing trade-offs between imperceptibility, robustness, capacity, and security. To meet these demands, deep learning-based fragile and semi-fragile methods, particularly those using INNs, emerge as the most promising due to their adaptability and resilience.
Based on this analysis, we provided practical guidelines to support the design and deployment of watermarking and steganographic methods in ICAO-compliant systems, identifying key features for effective integrity verification.
Despite promising advances, the field remains underexplored and requires further validation. Future research should evaluate the interaction between embedded signals and face recognition pipelines, assess resilience against adversarial and semantic attacks, and develop frameworks to verify compliance with ICAO standards. These steps will help bridge the gap between integrity verification and deployable security solutions in biometric identity management.

\section*{Acknowledgment}
This work was partially supported by Project SERICS (PE00000014) under the NRRP MUR program funded by the EU - NGEU. Davide Ghiani's PhD grant is partly funded by Dedem SpA under the PNRR program.

\clearpage

{\small
\bibliographystyle{ieee}
\bibliography{egbib}
}

\end{document}